\documentclass[runningheads]{llncs}
\usepackage{graphicx}
\usepackage{enumitem}
\usepackage{amsmath}
\usepackage{multibib}
\begin{document}
\title{Conflict Detection and Resolution in Table Top Scenarios for Human-Robot Interaction\thanks{We gratefully acknowledge funding by the Kempe Foundation under grant SMK-1644.}}

\author{Avinash Kumar Singh \and Kai-Florian Richter}

\authorrunning{A.K. Singh \& K.-F. Richter}
\titlerunning{Conflict resolution in human-robot interaction}
\institute{Department of Computing Science, Umeå University, Sweden
\email{\{avinash,kaifr\}@@cs.umu.se}}
\maketitle              % typeset the header of the contribution
\begin{abstract}
As in any interaction process, misunderstandings, ambiguity, and failures to correctly understand the interaction partner are bound to happen in human-robot interaction. We term these failures 'conflicts' and are interested in both conflict detection and conflict resolution. In that, we focus on the robot's perspective. For the robot, conflicts may occur because of errors in its perceptual processes or because of ambiguity stemming from human input. This poster presents a brief system overview, and details  Here, we briefly outline the project's motivation and setting, introduce the general processing framework, and then present two kinds of conflicts in some more detail: 1) a failure to identify a relevant object at all; 2) ambiguity emerging from multiple matches in scene perception. 

\end{abstract}

\section{Motivation, Scenario, and Approach}
Knowing when full autonomy will fail and collaboration with others is needed to successfully execute a task is a fundamental ability for humans to ensure efficiency, safety, and even survival. This ability is equally important for artificial cognitive agents, such as service and household robots or self-driving vehicles, who operate in our public or private spaces where they will often be faced with ill-defined or ambiguous human requests. Without this ability, these systems may get lost in their operations, in particular because these agents do not operate in isolation, but usually interact with others; and often with humans. Such interactions pose several challenges, in particular if the human interaction partner is `naive' with respect to the system's capabilities and inner workings--a situation that will be the norm once such agents will become part of our everyday lives.

The success of their introduction will depend as much on users' trust and willingness to cooperate as it will on the systems' technological and engineering capabilities. For example, as in human-human interaction misunderstandings and confusion are bound to happen. Thus, the systems' ability to cope with misunderstandings, ambiguity, and errors (termed `conflicts' here) in both perceptual processes and interaction with a human user will be highly important.

In our project we explore conflict detection and conflict resolution strategies for social robots. We use a simple scenario to focus on principle problems. Human and robot verbally interact on a table-top setting, where several small objects are placed in some arrangement on a table (e.g., cups, plates, cutlery; books, phones, laptops; fruits; or other typical small household items). The human would mention one of these objects, possibly further specifying its location using a referring expression~\cite{baranwal2019extracting}. For example, the human may say something like ``give me the cup'' or ``the book is next to the teapot.'' The robot would  parse this referring expression for the target object and any potentially mentioned objects and relations that further specify its location. It then would match that object to those identified in the object recognition step, similar to~\cite{shridhar2018interactive}. 

\begin{figure}[ht]
\centering
\includegraphics[width=.7\textwidth]{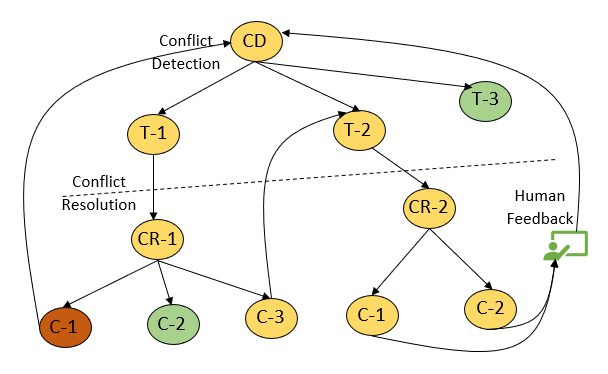}
\caption{Robot controller for conflict resolution (Conflict Detection (CD), Conflict Type-1,2,3 (T-1,2,3), Conflict Resolution (CR), Case-1,2,3 (C-1,2,3))} \label{fig1}
\end{figure}

In such a scenario, conflicts may arise because of issues correctly parsing human utterances (which we will not further address in this paper), because the instructions are ambiguous (to the robot at least), or because there is a mismatch between human instructions (or intentions) and the robot's scene interpretation. In other words, the robot fails to correctly identify the object intended by the human. In these cases, the robot a) needs to be able to identify this conflict, i.e., realize that it cannot (unambiguously) find the mentioned object and, b) have some strategies available to resolve such conflicts. For example, the robot may try to update its visual scene understanding by taking corrective actions, or it may go back to the human asking for clarification or more information. We term issue a) conflict detection (CD), and issue b) conflict resolution (CR). Both issues are further illustrated in the control flow diagram in Figure~\ref{fig1}.

Generally speaking, the matching of human referring expression to visual scene understanding may have several outcomes, further shown in Figure~\ref{fig1}. The parent node (Conflict Detection: CD) of the flow diagram represents the main controller which evaluates the referring expression with respect to recognized objects in the visual perception. If a unique match is found, there is no conflict (T-3)--at least none the robot could detect. Otherwise there is a conflict that needs to be resolved; the appropriate resolution strategy depends on the kind of conflict. The robot may either fail to recognize the mentioned object at all (T-1), or there are multiple detected objects that match the human object description (T-2).
To resolve conflicts of type 1 (CR-1), the robot would first increase the priors for the type of object searched for, i.e., raise the probability that such an object is present in the scene. One way of doing this is to lower the threshold for these objects in scene perception. Such thresholds introduce a minimum probability (certainty) in object recognition in order to avoid false positives, but sometimes they may also cause false negatives. Scanning the scene again with these lower thresholds may resolve the conflict (C-2), i.e., a unique match is found, or it may now find multiple matches (C-3). If the robot still does not detect any matches, it may change its perspective (view angle) on the scene by moving around and then restart the recognition process (CD). In case of a conflict of type 2 (T-2), the robot needs to resolve some ambiguity. To this end, it tries to identify those attributes of the objects that may disambiguate them (e.g., color, size, or shape; C-1) and initiate a clarification with the human using these attributes (e.g., `do you mean the blue or the red cup?'). If there are no suitable object attributes, it may use spatial (location) attributes instead (e.g., `The cup left of the banana or the one behind the book?'; C-2).

\section{System Implementation}
\label{system_implementation}
\begin{figure}[ht]
\centering
\includegraphics[width=.7\textwidth]{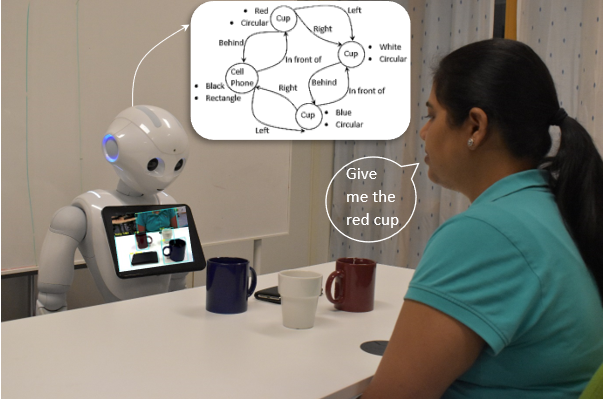}
\caption{Experimental setup: A human gives verbal instructions to the Pepper robot, which displays the detected objects on the tablet. Internally, it constructs a knowledge graph representing the perceived scene in front of it (shown in the box).} \label{fig2}
\end{figure}

We implemented a framework for scene recognition of table-top settings as described above. The framework runs on a Pepper robot and comprises of several different components. These include calibration of Pepper's two cameras (RGB-D), object detection (including color identification and shape estimation), extraction of spatial relations between objects, the construction of a knowledge graph representing the perceived scene, and language parsing. Figure~\ref{fig2} illustrates some of these components depicting the experimental setup.

Aligning both the robot's RGB and depth camera allows for identifying and representing objects in a three-dimensional space relative to the camera system, along with their attributes (e.g., color, shape, size). We use homography to this end.
In order to detect objects and to get their outer boundary we use a pre-trained Mask R-CNN model~\cite{he2017mask}.  Further, we have trained a Convolutional Neural Network (CNN) on a data-set of Google color images to recognize the color of objects (black, blue, green, orange, pink, purple, red, white, yellow). Spatial relations are extracted using a fuzzy inference system. We opt for fuzzy relations to cover the uncertainty inherent in processing visual scenes, but also in order to being able to capture differences in how humans may describe these relationships. Shape features are extracted using another fuzzy inference system~\cite{FES}. Language parsing is further described in~\cite{baranwal2019extracting} and allows for identifying the primary object, its attributes and relation with other objects. All this information is then fed into the controller (CD) as input to detect and resolve conflicts.

\section{Implications and Outlook}
The next steps in the project include finishing the implementation of an initial set of strategies for resolving the kinds of conflicts described above and then to run human-subject experiments for evaluating these. Experimental results will provide insights into how successful the strategies are in resolving conflicts, but also how participants perceive their use, i.e., whether they deem them beneficial in human-robot interaction.
As a wider implication, the scope and complexity in this project is deliberately limited and well-contained. Moving to more `open' worlds, for example, spaces that are no longer fully perceivable with a single camera view, or allowing for dynamics in the scene, will most likely create new sources of conflict, and offer new strategies for resolving these. Still, we believe the general framework designed in this project would continue to be valid.

\par Our further research in the arena of human-robot interaction can be seen in reference \cite{singh2014face-1}\cite{singh2012face}\cite{singh2016nao}\cite{singh2015sketch}\cite{singh2017development}\cite{baranwal2017development}\cite{singh2014face-2}\cite{baranwal2017real}\cite{singh2014expression}\cite{singh2019rough}\cite{singh2017visual}\cite{singh2016development}\cite{singh2015human}\cite{singh2019empirical}\cite{tripathy2013mfcc}\cite{baranwal2019fusion}
\cite{kumar2019towards}\cite{baranwal2017mathematical}\cite{baranwal2017efficient}\cite{baranwal2015possibility}\cite{tripathi2015continuous}\\
\cite{singh2014implementation}\cite{baranwal2014speaker}\cite{baranwal2014implementation}\cite{baranwal2014speech}\cite{tripathy2013mfcc}

\bibliography{References}

\begin{thebibliography}{10}

\bibitem{baranwal2014speech}
Neha Baranwal, Ganesh Jaiswal, and Gora~Chand Nandi.
\newblock A speech recognition technique using mfcc with dwt in isolated hindi
  words.
\newblock In {\em Intelligent Computing, Networking, and Informatics}, pages
  697--703. Springer, 2014.

\bibitem{baranwal2017efficient}
Neha Baranwal and Gora~Chand Nandi.
\newblock An efficient gesture based humanoid learning using wavelet descriptor
  and mfcc techniques.
\newblock {\em International Journal of Machine Learning and Cybernetics},
  8(4):1369--1388, 2017.

\bibitem{baranwal2017mathematical}
Neha Baranwal and Gora~Chand Nandi.
\newblock A mathematical framework for possibility theory-based hidden markov
  model.
\newblock {\em International Journal of Bio-Inspired Computation},
  10(4):239--247, 2017.

\bibitem{baranwal2017real}
Neha Baranwal, Gora~Chand Nandi, and Avinash~Kumar Singh.
\newblock Real-time gesture--based communication using possibility
  theory--based hidden markov model.
\newblock {\em Computational Intelligence}, 33(4):843--862, 2017.

\bibitem{baranwal2019extracting}
Neha Baranwal, Avinash~Kumar Singh, and Suna Bench.
\newblock Extracting primary objects and spatial relations from sentences.
\newblock In {\em 11th International Conference on Agents and Artificial
  Intelligence}, Prague, Czech Republic, 2019.

\bibitem{baranwal2019fusion}
Neha Baranwal, Avinash~Kumar Singh, and Thomas Hellstr{\"o}m.
\newblock Fusion of gesture and speech for increased accuracy in human robot
  interaction.
\newblock In {\em 2019 24th International Conference on Methods and Models in
  Automation and Robotics (MMAR)}, pages 139--144. IEEE, 2019.

\bibitem{baranwal2017development}
Neha Baranwal, Avinash~Kumar Singh, and Gora~Chand Nandi.
\newblock Development of a framework for human--robot interactions with indian
  sign language using possibility theory.
\newblock {\em International Journal of Social Robotics}, 9(4):563--574, 2017.

\bibitem{baranwal2014implementation}
Neha Baranwal, Neha Singh, and Gora~Chand Nandi.
\newblock Implementation of mfcc based hand gesture recognition on hoap-2 using
  webots platform.
\newblock In {\em 2014 International Conference on Advances in Computing,
  Communications and Informatics (ICACCI)}, pages 1897--1902. IEEE, 2014.

\bibitem{baranwal2015possibility}
Neha Baranwal, Kumud Tripathi, and GC~Nandi.
\newblock Possibility theory based continuous indian sign language gesture
  recognition.
\newblock In {\em TENCON 2015-2015 IEEE Region 10 Conference}, pages 1--5.
  IEEE, 2015.

\bibitem{baranwal2014speaker}
Neha Baranwal, Shweta Tripathi, and Gora~Chand Nandi.
\newblock A speaker invariant speech recognition technique using hfcc features
  in isolated hindi words.
\newblock {\em International Journal of Computational Intelligence Studies},
  3(4):277--291, 2014.

\bibitem{he2017mask}
Kaiming He, Georgia Gkioxari, Piotr Doll{\'a}r, and Ross Girshick.
\newblock Mask {R-CNN}.
\newblock In {\em Proceedings of the IEEE International Conference on Computer
  Vision}, pages 2961--2969, 2017.

\bibitem{kumar2019towards}
Avinash Kumar~Singh, Neha Baranwal, Kai-Florian Richter, Thomas Hellstr{\"o}m,
  and Suna Bensch.
\newblock Towards verbal explanations by collaborating robot teams.
\newblock In {\em International Conference on Social Robotics (ICSR’19),
  Workshop Quality of Interaction in Socially Assistive Robots, Madrid, Spain,
  November 26-29, 2019.}, 2019.

\bibitem{shridhar2018interactive}
Mohit Shridhar and David Hsu.
\newblock Interactive visual grounding of referring expressions for human-robot
  interaction.
\newblock {\em arXiv preprint:1806.03831}, 2018.

\bibitem{singh2015human}
Avinash~Kumar Singh, Neha Baranwal, and Gora~Chand Nandi.
\newblock Human perception based criminal identification through human robot
  interaction.
\newblock In {\em 2015 Eighth International Conference on Contemporary
  Computing (IC3)}, pages 196--201. IEEE, 2015.

\bibitem{singh2017development}
Avinash~Kumar Singh, Neha Baranwal, and Gora~Chand Nandi.
\newblock Development of a self reliant humanoid robot for sketch drawing.
\newblock {\em Multimedia Tools and Applications}, 76(18):18847--18870, 2017.

\bibitem{singh2019rough}
Avinash~Kumar Singh, Neha Baranwal, and Gora~Chand Nandi.
\newblock A rough set based reasoning approach for criminal identification.
\newblock {\em International Journal of Machine Learning and Cybernetics},
  10(3):413--431, 2019.

\bibitem{singh2019empirical}
Avinash~Kumar Singh, Neha Baranwal, and Kai-Florian Richter.
\newblock An empirical review of calibration techniques for the pepper humanoid
  robot’s rgb and depth camera.
\newblock In {\em Proceedings of SAI Intelligent Systems Conference}, pages
  1026--1038. Springer, 2019.

\bibitem{singh2015sketch}
Avinash~Kumar Singh, Pavan Chakraborty, and GC~Nandi.
\newblock Sketch drawing by nao humanoid robot.
\newblock In {\em TENCON 2015-2015 IEEE Region 10 Conference}, pages 1--6.
  IEEE, 2015.

\bibitem{singh2014face-2}
Avinash~Kumar Singh, Piyush Joshi, and Gora~Chand Nandi.
\newblock Face liveness detection through face structure analysis.
\newblock {\em International Journal of Applied Pattern Recognition},
  1(4):338--360, 2014.

\bibitem{singh2014face-1}
Avinash~Kumar Singh, Piyush Joshi, and Gora~Chand Nandi.
\newblock Face recognition with liveness detection using eye and mouth
  movement.
\newblock In {\em 2014 International Conference on Signal Propagation and
  Computer Technology (ICSPCT 2014)}, pages 592--597. IEEE, 2014.

\bibitem{singh2016development}
Avinash~Kumar Singh, Piyush Joshi, and Gora~Chand Nandi.
\newblock Development of a fuzzy expert system based liveliness detection
  scheme for biometric authentication.
\newblock {\em arXiv preprint arXiv:1609.05296}, 2016.

\bibitem{singh2014expression}
Avinash~Kumar Singh, Arun Kumar, GC~Nandi, and Pavan Chakroborty.
\newblock Expression invariant fragmented face recognition.
\newblock In {\em 2014 International Conference on Signal Propagation and
  Computer Technology (ICSPCT 2014)}, pages 184--189. IEEE, 2014.

\bibitem{singh2012face}
Avinash~Kumar Singh and Gora~Chand Nandi.
\newblock Face recognition using facial symmetry.
\newblock In {\em Proceedings of the Second International Conference on
  Computational Science, Engineering and Information Technology}, pages
  550--554. ACM, 2012.

\bibitem{singh2016nao}
Avinash~Kumar Singh and Gora~Chand Nandi.
\newblock Nao humanoid robot: Analysis of calibration techniques for robot
  sketch drawing.
\newblock {\em Robotics and Autonomous Systems}, 79:108--121, 2016.

\bibitem{singh2017visual}
Avinash~Kumar Singh and Gora~Chand Nandi.
\newblock Visual perception-based criminal identification: a query-based
  approach.
\newblock {\em Journal of Experimental \& Theoretical Artificial Intelligence},
  29(1):175--196, 2017.

\bibitem{singh2014implementation}
Neha Singh, Neha Baranwal, and GC~Nandi.
\newblock Implementation and evaluation of dwt and mfcc based isl gesture
  recognition.
\newblock In {\em 2014 9th International Conference on Industrial and
  Information Systems (ICIIS)}, pages 1--7. IEEE, 2014.

\bibitem{tripathi2015continuous}
Kumud Tripathi, Neha Baranwal, and Gora~Chand Nandi.
\newblock Continuous dynamic indian sign language gesture recognition with
  invariant backgrounds.
\newblock In {\em 2015 International Conference on Advances in Computing,
  Communications and Informatics (ICACCI)}, pages 2211--2216. IEEE, 2015.

\bibitem{tripathy2013mfcc}
Shweta Tripathy, Neha Baranwal, and GC~Nandi.
\newblock A mfcc based hindi speech recognition technique using htk toolkit.
\newblock In {\em 2013 IEEE Second International Conference on Image
  Information Processing (ICIIP-2013)}, pages 539--544. IEEE, 2013.

\bibitem{FES}
Linghao Zhang.
\newblock Recognizing simple geometrical figures using fuzzy expert system.
\newblock \url{https://github.com/dnc1994/Shape/blob/master/report.pdf}.
\newblock Accessed: 2019-04-19.

\end{thebibliography}

\end{document}